\documentclass{article}
\usepackage{spconf}
\usepackage[cmex10]{amsmath}
\usepackage{amsthm}
\usepackage{amssymb}
\usepackage{mathrsfs}
\usepackage{graphicx}
\usepackage{float}
\usepackage{array}
\usepackage{epstopdf}
\usepackage{multirow}

\title{F\MakeLowercase{ingerprint} R\MakeLowercase{ecognition} U\MakeLowercase{sing} T\MakeLowercase{ranslation} I\MakeLowercase{nvariant} S\MakeLowercase{cattering} N\MakeLowercase{etwork}}
\name{}
\name{Shervin Minaee and Yao Wang}
\address{Electrical and Computer Engineering Department, New York University, USA.}
\allowdisplaybreaks[1]	
\begin{document}
%
\maketitle
\begin{abstract}
Fingerprint recognition has drawn a lot of attention during last decades. Different features and algorithms have been used for fingerprint recognition in the past.
In this paper, a powerful image representation called scattering transform/network, is used for recognition. Scattering network is a convolutional network where its architecture and filters are predefined wavelet transforms.
The first layer of scattering representation is similar to sift descriptors and the higher layers capture higher frequency content of the signal.
After extraction of scattering features, their dimensionality is reduced by applying principal component analysis (PCA). At the end, multi-class SVM is used to perform template matching for the recognition task.
The proposed scheme is tested on a well-known fingerprint database and has shown promising results with the best accuracy rate of 98\%.
\end{abstract}

\section{Introduction}
\label{sec:intro}
To make an application more secure and less accessible to undesired people, we need to be able to distinguish a person from the others. There are various ways to identify a person such as keys, passwords and cards. However,  biometrics are the most secure options so far. They are virtually impossible to imitate by any other than the desired person himself. They can be divided into two categories: behavioral and physiological features. Behavioral features are those actions that a person can uniquely create or express, such as signatures, walking rhythm, and the physiological features are those characteristics that a person possesses, such as fingerprints and iris pattern. Many works revolved around recognition and categorization of such data including, but not limited to, fingerprints, faces, palmprints and iris patterns \cite{fing_book}-\cite{Iris}. 

Fingerprint is perhaps one of the most popular biometrics. It has been used in various applications such as forensics, transaction authentication, etc \cite{fing_book}. 
Many of the algorithms proposed for fingerprint recognition are minutiae-based matching.
The major minutiae features of fingerprint ridges are ridge ending, bifurcation, and short ridge.
In many of these algorithms, minutiae are extracted from the test and input fingerprint images, and the number of corresponding minutiae pairings between these two images is used to verify the test fingerprint image. 
In the case of low quality fingerprint images, new foreground segmentation approaches can be used to extract the minutiae from fingerprints with an enhanced quality \cite{LAD_seg}.
There are also a lot of image representations and feature-based algorithms for fingerprint recognition.
In \cite{SIFT_fing}, Park proposed a fingerprint recognition system based on SIFT features. They extract SIFT feature points in scale space and perform matching based on the texture information around the feature points using the SIFT operator.
Among more recent works, in \cite{Cappelli}, Cappelli proposed a new representation based on 3D data structure built from minutiae distances and angles called Minutia Cylinder-Code (MCC). In \cite{DP}, Zhao proposed to use pore matching approach toward fingerprint recognition. In \cite{ADM}, Zhao proposed an adaptive pore modeling for fingerprint recognition.

Many of the biometric recognition systems involve a lot of pre-processing steps which are specifically designed for that kind of data and the final performance largely depends on the goodness of those steps. Most of them use a single layer representation of the image which may not be able to extract very discriminative set of features, and some of them may work well on some of the datasets but not on the other ones. 
Therefore, there have been a lot of efforts to design some supervised or unsupervised feature representation which works pretty well over various datasets and problems. These image representations should have invariance with respect to intra-class variation in the data. 
Scattering transform/network is one of such representations.
Scattering network is a convolutional network in which the filters and architectures are predefined wavelet filters \cite{scat3}.
It can be designed such that it is invariant to a family of transformation and small deformations \cite{scat4}. Due to tremendous success of deep scattering networks to achieve state-of-the-art results in several image and audio classification benchmarks \cite{scat1}, it is interesting to know how this representation works for biometric recognition. 
In \cite{my_iris}, the scattering transform is used for iris recognition and achieved very high accuracy rate. 
It can also be for extraction of features from MRI and other medical images \cite{mri1}. 
Here scattering transformation is used for fingerprint recognition. One advantage of deep scattering network is that all the architecture is known in advance and it does not require any learning of the weights and one could get very rich set of features by going up to two levels in this network.
Therefore the proposed algorithm is very fast and can be implemented in electronic devices in conjunction with energy-efficient algorithms \cite{hoseini1}, \cite{hoseini2}.
After the scattering features are extracted, their dimensionality is reduced using PCA \cite{PCA}. At the end, multi-class SVM is used to perform classification using PCA features. This algorithm is tested on the well-known PolyU fingerprint database \cite{polyu} and achieved very high accuracy rate.
Four sample fingerprint images of this database are shown in Figure 1.
\begin{figure}[1 h]
\begin{center}
    \includegraphics [scale=0.44] {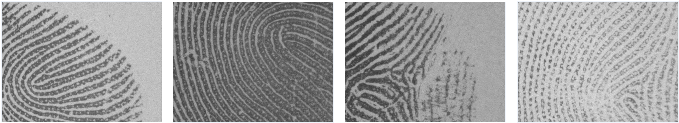}
\end{center}
\vspace{-0.2cm}
  \caption{Four different fingerprint images}
\end{figure}

The rest of this paper is organized as follows. Section 2 describes the features which are used in this work. The details of scattering transformation is provided in Section 2.1 and the PCA algorithm is explained in Section 2.2. Section 3 contains the explanation of the classification scheme. The results of our experiments and comparisons with other works are presented in Section 4 and the paper is concluded in Section 5.

\section{Features}
\label{SectionII}
Images of the same object could have variablity due to translation, scale, rotation, illumination changes. These changes in the images of a single object class are called intra-class variations which make object recognition very difficult in some scenarios. Therefore it is very important to design some image representations which are invariant to some of these intra-class variations. 
Various image descriptors have been proposed during past 20 years. SIFT and HOG are two popular hand-crafted image descriptors which achieved very good results on several object recognition tasks. Sparse representation has also been used for extracting features in image classification task \cite{hojat1}-\cite{hojat2}.
But some of these traditional descriptors are not very successful for some of the more challenging datasets with many object classes and large intra-class variations. In the more recent works, deep neural network and also dictionary learning approaches have achieved state-of-the-art results on various datasets, most notably Alex-net \cite{Alex} which is trained on ImageNet competition. In deep learning framework, the images are fed as the input the multi-layer neural network and the network itself figures out what is the best way to combine the pixels for maximizing the accuracy.
In the dictionary learning approach, different algorithms such as K-SVD or K-LDA are used to learn a set of features which are suitable for a given training set \cite{ksvd}, \cite{klda}.
In a recent work, a wavelet-based representation is proposed by Mallat \cite{scat1}, which is similar to deep convolutional network where instead of learning the filters and representation, it uses predefined wavelets \cite{scat2}. These wavelets can be adopted such that they achieve some desired geometric invariance such as translation, rotation and scale invariance \cite{scat3}.
The details of scattering transformation are described in the following section.

\subsection{Scattering Features}
The scattering operator is a deep convolutional network which uses wavelet transform as its filter and can be designed such it is invariant to group of transformations such as translation, rotation, etc \cite{scat3}. The scattering transform computes local image descriptors with a cascade of three operations: wavelet decompositions, complex modulus and a local averaging. The scattering transform provides a multi-layer representations for a signal.
As discussed in \cite{scat1}, some other image descriptors such as SIFT can be obtained by averaging the amplitude of wavelet coefficients, calculated using directional wavelets. This averaging provides translation-invariance to some extent, but it also reduces the high-frequency information. Scattering transform is designed such that it recovers the high-frequency information lost by this averaging. It can be shown that the coefficients in the first layer of the scattering transform are similar to SIFT descriptors and the coefficients in the higher layers contain higher-frequency information of the image.

We can get different versions of scattering transform by modifying it such that it is invariant to a new family of transformations. In this work, translation invariant scattering transform is used and a brief description of that is provided here.

Suppose we have a signal $f(x)$. The first scattering coefficient is the average of the signal and can be obtained by convolving the signal with an averaging filter $\phi_J$ as $f*\phi_J$. The scattering coefficients of the first layer can be obtained by applying wavelet transforms at different scales and orientations, removing the complex phase and taking their average by $\phi_J$ as shown below:
\begin{gather*}
|f* \psi_{j_1,\lambda_1}|*\phi_J
\end{gather*}
where $j_1$ and $\lambda_1$ denote different scales and orientations. Taking the magnitude of the wavelet coefficients can be thought of the non-linear pooling functions used in convolutional neural networks.
Note that by removing the complex phase of wavelet we can make these coefficients insensitive to local translation.

Now to recover the high-frequency contents of the signal, which are eliminated from the wavelet coefficients of first layer by averaging, we can convolve the $|f* \psi_{j_1,\lambda_1}|$ by another set of wavelet at scale $j_2<J$, taking the absolute value of wavelet and taking the average:
\begin{gather*}
||f* \psi_{j_1,\lambda_1}|*\psi_{j_2,\lambda_2}|*\phi_J
\end{gather*}
It can be shown that $|f* \psi_{j_1,\lambda_1}|*\psi_{j_2,\lambda_2}$ is negligible for scales where $2^{j_1} \leq  2^{j_2}$. Therefore we only need to calculate the coefficients for $j_1 >j_2 $.

The convolution with $\phi_J$ at the second layer removes high frequencies and results in locally translation-invariant second-order coefficients. This high-frequency information can be restored again by finer scale wavelet coefficients in the next layers. 
We can continue this procedure to obtain the coefficients of the $k$-{th} layer of scattering network as:
\begin{gather*}
\underset{\ \ \ \ \ \ \ \ \ \ \ \ \ \ \ \ \ \ \ \ \ \ \ \ \ \ j_k<...<j_2<j_1<J, \ (\lambda_1,...,\lambda_k) \in \Gamma^k  }{S_{k,J}(f(x)))= ||f* \psi_{j_1,\lambda_1}|*...*\psi_{j_k,\lambda_k}|*\phi_J}
\end{gather*}

It can be shown that the scattering vector of the $k$-{th} layer has a size of $p^k {J \choose k}$ where $p$ denotes the number of different orientations and $J$ denotes the number of scales.
A scattering vector is formed as the concatenation of the coefficients of all layers up to $m$ which has a size of $\sum_{k=0}^m{p^k {J \choose k}}$. For many signal processing applications, a scattering network with two or three layers will be enough.
At the end, we can extract the mean and variance from each scattering transform image to form the scattering feature vector. One can also extract further information from each image to form the scattering feature vector.

The transformed images of the first and second layers of scattering transform for a sample fingerprint image are shown in Figures 2 and 3. These images are derived by applying bank of filters of 5 different scales and 6 orientations.

\begin{figure}[2 h]
\begin{center}
    \includegraphics [scale=0.35] {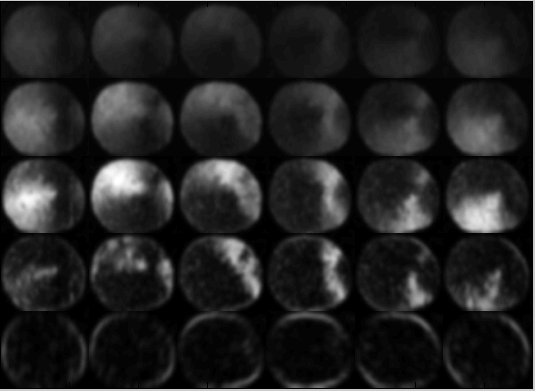}
\end{center}
  \vspace{-0.2cm}
  \caption{The images from the first layer of scattering transform}
\end{figure}
\begin{figure}[3 h]
\begin{center}
    \includegraphics [scale=0.42] {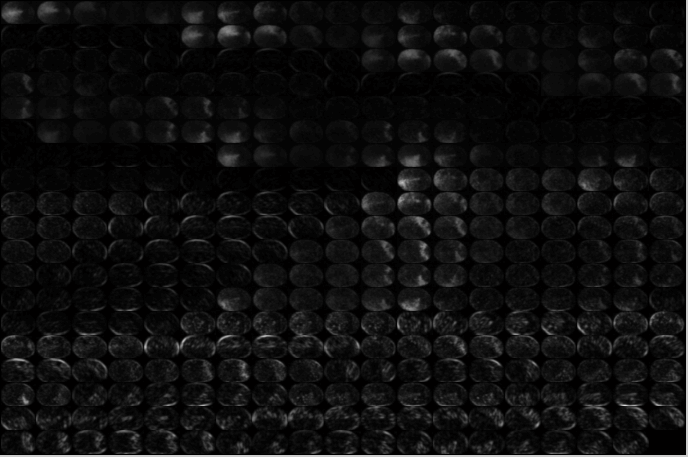}
\end{center}
\vspace{-0.2cm}
  \caption{The images from the second layer of scattering transform}
\end{figure}

\subsection{Principal Component Analysis}
In a lot of applications, one need to reduce the dimensionality of the data to make the algorithm faster and more efficient. Principal component analysis (PCA) is a powerful algorithm used for dimensionality reduction \cite{PCA}. Given a set of correlated variables, PCA transforms them into another domain such the transformed variables are linearly uncorrelated. This set of linearly uncorrelated variables are called principal components. PCA is usually defined in a way that the first principal component has the largest possible variance, the second one has the second largest variance and so on. Therefore after applying PCA, we could only keep a subset of principal components with the largest variance to reduce the dimensionality.
PCA has a lot of applications in computer vision and neuroscience. Eigenface is one representative application of PCA in computer vision, where PCA is used for face recognition.

Let us assume we have a dataset of $M$ fingerprint images and $\{f_1,f_2,...,f_M\}$ denote their features where $f_i \in \mathbf{R}^d$. To apply PCA, all features need to be centered first by subtracting their mean:  $z_i= f_i- \bar{f}$ where $\bar{f}= \frac{1}{M} \sum_{i=1}^M f_i$.
Then the covariance matrix of the centered images is calculated as:
\begin{gather*}
C= \sum_{i=1}^M z_i z_i^T 
\end{gather*}
Next the eigenvalues $\lambda_j$ and eigenvectors $\nu_j$ of the covariance matrix $C$ are computed. Suppose $\lambda_j$'s are ordered based on their values. Then each $z_i$ can be written as $z_i= \sum_{j=1}^d \alpha_j \nu_j$.
The dimensionality of the data can be reduced by projecting them on the first $K (\ll d)$ principal vectors as:
\begin{gather*}
\hat{z_i}= (\nu_1^{T} z_i, \nu_2^{T} z_i,..., \nu_K^{T} z_i)= (\alpha_1,...,\alpha_K)
\end{gather*}

By keeping $k$ principal components, the percentage of retained variance can be found as $\frac{\sum_{j=1}^k \lambda_j}{\sum_{j=1}^d \lambda_j }$. 
One issue is how to choose the value $k$, the number of principal components. One simple way to choose $k$ would be to pick a value such that the above ratio is less than $\epsilon$, where $\epsilon$ is usually chosen between 95\% to 99\%.

\section{Recognition Algorithm: Support Vector Machine}
\label{SectionIII}
After capturing the features of all people in the dataset, a classifier should be used to find the closest match of each test sample. There are various classifiers which can be used for this task including support vector machine (SVM) \cite{SVM}, majority voting algorithm and neural network. In this work multi-class SVM has been used which is quite popular for image classification. 
A brief overview of SVM for binary classification is presented here. For further detail and extensions to multi-class settings we refer the reader to \cite{multi_SVM}.
Let us assume we want to separate the set of training data $(x_1,y_1)$, $(x_2,y_2)$, ..., $(x_n,y_n)$ into two classes where 
$x_i \in \mathbf{R}^d $ is the feature vector and $y_i \in \{-1,+1\}$ is the class label. If we assume two classes are linearly separable with a hyperplane $w.x+b=0$ with no other prior knowledge about the data, then the optimal hyperplane is the one with the maximum margin. One can show that the maximum margin hyperplane can be found by the following optimization problem:
\begin{equation}
\begin{aligned}
& \underset{w,b}{\text{minimize}}
& & \frac{1}{2} ||w||^2 \\
& \text{subject to}
& & y_i(w.x_i+b) \geq 1, \; i = 1, \ldots, n.
\end{aligned}
\end{equation} \\
Since this problem is convex, we can solve it by looking at the dual problem and introducing Lagrange multipliers $\alpha_i$ which results in the following classifier:
\begin{equation}
f(x)= sign(\sum_{i=1}^{n} \alpha_i y_i w.x+b)
\end{equation}
$\alpha_i$ and $b$ are calculated by the SVM learning algorithm. Interestingly, after solving the dual optimization problem, most of the $\alpha_i$'s are zero; those datapoints $x_i$ which have nonzero $\alpha_i$ are called support-vectors.
There is also a soft-margin version of SVM which allows for mislabeled examples. If there exists no hyperplane that can split the "-1" and "+1" examples, the soft-margin method will choose a hyperplane that splits the examples as cleanly as possible, while still maximizing the distance to the nearest cleanly split examples \cite{SVM}. It introduces some penalty term in the primal optimization problem with misclassification penalty of $C$ times the degree of misclassification.

To derive the nonlinear classifier, one can map the data from input space into a higher-dimensional feature space $\mathcal{H}$ as: $x\rightarrow \phi(x)$, so that the classes are linearly separable in the feature space \cite{kernel_SVM}. If we assume there exists a kernel function where $k(x,y)= \phi(x).\phi(y)$, then we can use the kernel trick to construct nonlinear SVM by replacing the inner product $x.y$ with $k(x,y)$ which results in the following classifier:
\begin{equation}
f_n(x)= sign(\sum_{i=1}^{n} \alpha_i y_i K(x,x_i)+b)
\end{equation}

To derive multi-class SVM for a set of data with $M$ classes, we can train $M$ binary classifiers which can discriminate each class against all other classes, and to choose the class which classifies the test sample with greatest margin   (one-vs-all). 
In another approach, we can train  a set of $M \choose 2$ binary classifiers which any of them separates one class from another one and to choose the class that is selected by the most classifiers. There are also some other approaches for multi-class SVM.

\section{Experimental results and analysis}
\label{SectionIV}
A detailed description of experimental results is presented in this section.
First, let us describe the parameter values of our algorithm.
For each image, scattering transform is applied up to two levels with a set of filter banks with 5 scales and 6 orientations, resulting in 391 transformed images. From each image the mean and variance are calculated and used as features, resulting in 782 scattering features. For scattering transformation, we used the software implemented by Mallat's group \cite{scat}. 
Then PCA is applied to all features and the first 200 PCA features are used for recognition. Multi-class SVM is used for the template matching. For SVM, we have used LIBSVM library \cite{libsvm}, and linear kernel is used with the penalty cost $C=1$.

We have tested our algorithm on the PolyU fingerprint database which is provided by Hong Kong Polytechnic University. It contains 1480 images of 148 fingers. The images of 25 people are used as a validation set for parameter tuning of our algorithm.  Then from the remaining fingers, half of the images are used for training and the other half for testing. To make feature extraction faster, we have resized all images to $80 \times 60$.

Figure 4 shows the recognition rate of the proposed approach for different number of PCA features. 
Interestingly, even by using few PCA features, we are able to get a very high accuracy rate.
As it can be seen, using 200 PCA features results in an accuracy rate around 98\%, which will not increase much by using more PCA features.

\begin{figure}[4 h]
\begin{center}
    \includegraphics [scale=0.4] {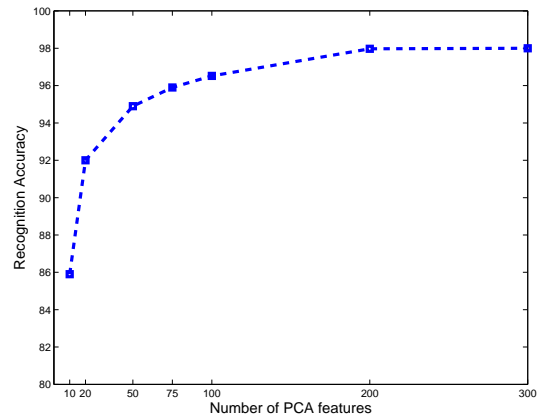}
\end{center}
  \vspace{-0.1cm}
  \caption{Recognition accuracy as a function of number of PCA features}
\end{figure}

The equal error rate (EER) of the proposed algorithm is also calculated on this dataset.
Equal error rate is a rate at which both acceptance and rejection errors are equal. 
To find EER we have used the minimum distance classifier. Figure 5 shows the false acceptance rate and false rejection rate versus the distance threshold. As we can see using the proposed the EER= 8\% is achieved.
 
\begin{figure}[5 h]
\begin{center}
    \includegraphics [scale=0.4] {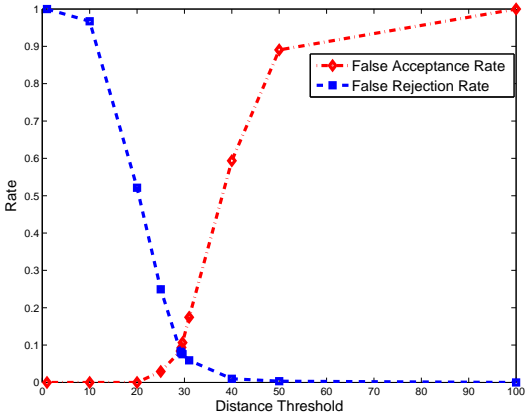}
\end{center}
  \vspace{-0.1cm}
  \caption{FAR and FRR versus the distance threshold}
\end{figure}

Table 1 shows a comparison between the EER of the proposed scheme and those of some other previous works on this dataset. The proposed approach achieved a smaller EER compared to the other approaches, but there is still a big room for improvement of EER on this dataset. That shows that the translation-invariant scattering network is more suitable for fingerprint identification than verification.
The EER can be further improved by using rotation-translation invariant scattering features and also using more powerful classifiers than minimum distance classifier.

\begin{table} [h]
\centering
  \caption{A comparison between EER of the proposed scheme and previous approaches }
  \centering
\begin{tabular}{|m{4.8cm}|m{3cm}|}
\hline
\ \ \ \ \ \ \ \ \ \ \ \ \ \ \ \ \ \ \ \ Method & \ \  \ \ equal error rate\\
\hline
MICPP \cite{DP} & \ \ \ \ \ 30.45\% \\
\hline
Direct Pore Matching \cite{DP} & \ \ \ \ \  20.49\% \\
\hline
Using Only Minutiae \cite{DP} & \ \ \ \ \ 17.68\% \\
\hline
Adaptive pore modeling \cite{ADM} & \ \ \ \ \ 11.51\% \\
\hline
The proposed scheme & \ \ \ \ \  8.1\% \\
\hline
\end{tabular}
\label{TblComp}
\end{table}

The experiments are performed using MATLAB 2012 on a laptop with Core i5 CPU running at 2.6GHz. 
It takes around 97 milliseconds for each image to perform template matching using multi-class SVM.

\section{Conclusion}
\label{SectionV}
This paper proposed to use a translation-invariant scattering network for fingerprint recognition. Scattering features are locally invariant and carry a lot of high-frequency information which are lost in other descriptors such as SIFT. The high-frequency information provides great discriminating power for fingerprint recognition.
Then PCA is applied on features to reduce dimensionality. At the end, multi-class SVM is used to perform template matching.
This shows the potential of scattering network for biometric recognition systems. In the future, we will investigate to apply the proposed set of features to other biometrics.

\section*{Acknowledgments}
The authors would like to thank Stephane Mallat's research group at ENS for providing the software implementation of scattering transform. We would also like to thank the CSIE group at NTU for providing LIBSVM software.
We would also like to thank biometric research group at PolyU Hong Kong for providing the fingerprint dataset.

\end{document}